\documentclass[11pt,a4paper]{article}
\usepackage{authblk}
\usepackage[hyperref]{acl2019}
\usepackage{booktabs}
\usepackage{times}
\usepackage{latexsym}
\usepackage[utf8]{inputenc}
\usepackage{mathtools}
\usepackage{colortbl}
\usepackage{comment}
\usepackage{todonotes}
\usepackage{subcaption}
\usepackage{graphicx}
\usepackage{url}
\usepackage{verbatim}
\usepackage{dsfont}
\usepackage{amsmath,amsfonts,amssymb}

\aclfinalcopy 
\def\aclpaperid{1531} 

\title{ Every child should have parents: a taxonomy refinement algorithm \\ based on hyperbolic term embeddings}

\author[1]{\textbf{Rami Aly}}
\author[2]{\textbf{Shantanu Acharya}}
\author[1]{\textbf{Alexander Ossa}}
\author[4,1]{\textbf{Arne Köhn}}
\author[1]{\\\textbf{Chris Biemann}}
\author[3,1]{\textbf{Alexander Panchenko}}

\affil[1]{Universit{\"a}t  Hamburg, Hamburg, Germany}
\affil[2]{National Institute of Technology Mizoram, Aizawl, India}
\affil[3]{Skolkovo Institute of Science and Technology, Moscow, Russia}
\affil[4]{Saarland University, Saarbr\"ucken, Germany}

\affil[ ]{\href{mailto:panchenko@informatik.uni-hamburg.de}{\{5aly,2ossa,koehn,biemann,panchenko\}@informatik.uni-hamburg.de}}

\date{}

\begin{document}
\maketitle

\begin{abstract}
We introduce the use of Poincaré embeddings to improve existing state-of-the-art approaches to domain-specific taxonomy induction from text as a signal for both relocating wrong hyponym terms within a (pre-induced) taxonomy as well as for attaching disconnected terms in a taxonomy.
This method substantially improves previous state-of-the-art results on the SemEval-2016 Task 13 on taxonomy extraction. We demonstrate the superiority of Poincaré embeddings over distributional semantic representations, supporting the hypothesis that they can better capture hierarchical lexical-semantic relationships than embeddings in the Euclidean space.
\end{abstract}

\section{Introduction}

The task of taxonomy induction aims at creating a semantic hierarchy of entities by using hyponym-hypernym relations -- called \emph{taxonomy} -- from text corpora.
Compared to many other domains of natural language processing that make use of pre-trained dense representations,
state-of-the-art taxonomy learning is still highly relying on traditional approaches like extraction of lexical-syntactic patterns \citep{Hearst:1992:AAH:992133.992154} or co-occurrence information \citep{grefenstette2015inriasac}. 
Despite the success of pattern-based approaches, most taxonomy induction systems suffer from a significant number of disconnected terms, since the extracted relationships are too specific to cover most words \citep{wang2017short, bordea2016semeval}. The use of distributional semantics for hypernym identification and relation representation has thus received increasing attention
\cite{shwartz2016improving}. However,
\citet{levy2015supervised} observe that many proposed supervised approaches instead learn prototypical hypernyms (that are hypernyms to many other terms), not taking into account the relation between both terms in classification. 
Therefore, past applications of distributional semantics appear to be rather unsuitable to be directly applied to taxonomy induction as the sole signal~\cite{tan2015usaar, pocostales2016nuig}. 
We address that issue by introducing a series of simple and parameter-free refinement steps that employ word embeddings in order to improve existing domain-specific taxonomies, induced from text using traditional approaches in an unsupervised fashion. 

We compare two types of dense vector embeddings: the standard word2vec CBOW model \cite{cbow, mikolov2013distributed}, that embeds terms in \emph{Euclidean space} based on distributional similarity, and the more recent Poincaré embeddings \cite{NIPS2017_7213}, which capture similarity as well as hierarchical relationships in a \emph{hyperbolic space}. The source code has been published\footnote{\url{https://github.com/uhh-lt/Taxonomy_Refinement_Embeddings}} to recreate the employed embedding, to refine taxonomies as well as to enable further research of Poincaré embeddings for other semantic tasks.

\section{Related Work}
The extraction of taxonomic relationships from text corpora is a long-standing problem in ontology learning, see \citet{Biemann2005ontolog} for an earlier survey. \citet{wang2017short} discuss recent advancements in taxonomy construction from text corpora.
Conclusions from the survey include: i) The performance of extraction of IS-A relation can be improved by studying how pattern-based and distributional approaches complement each other; ii) there is only limited success of pure deep learning paradigms here, mostly because it is difficult to design a single objective function for this task. 

On the two recent TExEval tasks at SemEval for taxonomy extraction \citep{task17semeval2015,bordea2016semeval}, attracting a total of 10 participating teams, attempts to primarily use a distributional representation failed. This might seem counterintuitive, as taxonomies are surely modeling semantics and thus their extraction should benefit from semantic representations. The 2015 winner INRIASAC \cite{grefenstette2015inriasac}  performed relation discovery using substring inclusion, lexical-syntactic patterns and co-occurrence information based on sentences and documents from Wikipedia. The winner in 2016, TAXI \citep{panchenko2016taxi}, harvests hypernyms with substring inclusion and Hearst-style lexical-syntactic patterns \cite{Hearst:1992:AAH:992133.992154} from domain-specific texts obtained via focused web crawling.
The only submission to the TExEval 2016 task that relied exclusively on distributional semantics to induce hypernyms by adding a vector offset to the corresponding hyponym \cite{pocostales2016nuig}  achieved only modest results. A more refined approach to applying distributional semantics by \citet{Zhang:2018:TUT:3219819.3220064} generates a hierarchical clustering of terms with each node consisting
of several terms.  They find concepts that should stay in the same
cluster using embedding similarity -- whereas, similar to the TExEval task, we are interested in making distinctions between all terms. 
Finally, \citet{le2019inferring} also explore using Poincar\'e embeddings for taxonomy induction, evaluating their method on hypernymy detection and reconstructing WordNet. However, in contrast to our approach that filters and attaches terms, they perform inference.

\section{Taxonomy Refinement using Hyperbolic Word Embeddings}
\begin{figure*}
\centering
\includegraphics[width=1\textwidth, trim = {0, 0cm, 0, 0cm}, clip]{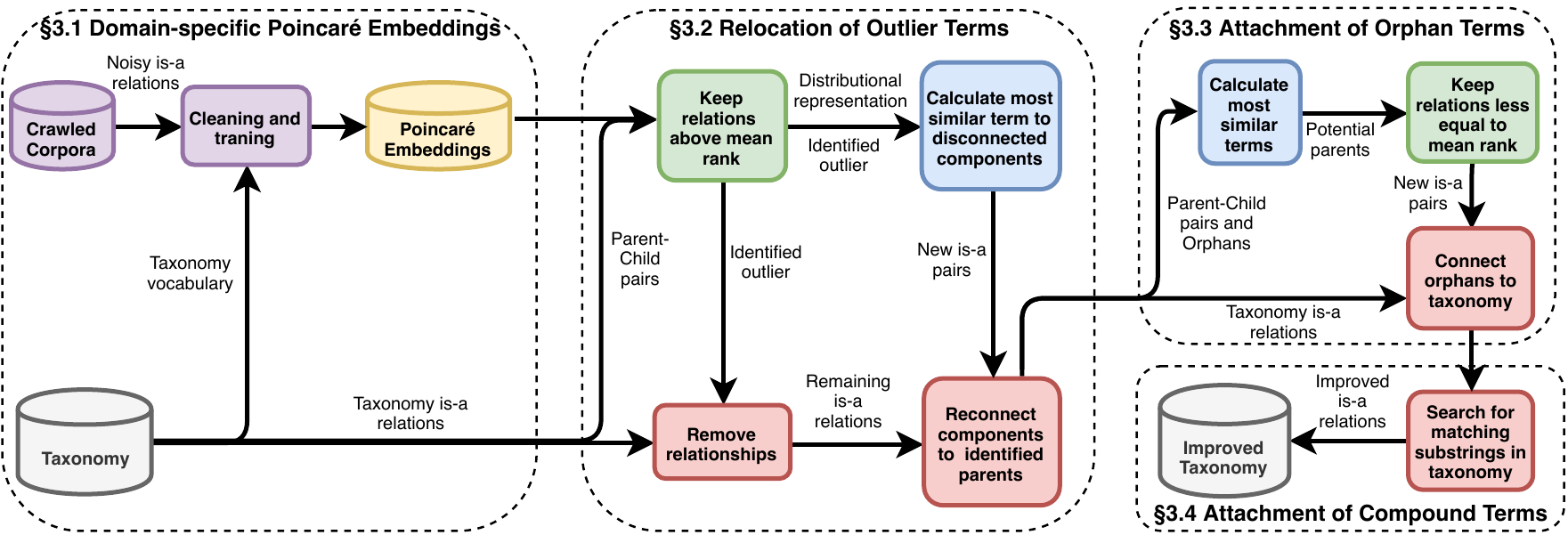}
\caption{Outline of our taxonomy refinement method, with paper sections indicated.}
\label{process_chart}
\vspace{-1em}
\end{figure*}

We employ embeddings using distributional semantics (i.e. word2vec CBOW) and Poincaré embeddings \cite{NIPS2017_7213} to alleviate the largest error classes in taxonomy extraction: the existence of \emph{orphans} -- disconnected nodes that have an overall connectivity degree of zero and \emph{outliers} -- a child node that is assigned to a wrong parent. The rare case in which multiple parents can be assigned to a node has been ignored in the proposed refinement system.
The first step consists of creating domain-specific Poincaré embeddings (§~\ref{Poincare_step}). They are then used to identify and relocate outlier terms in the taxonomy (§~\ref{outlier_replace}), as well as to attach unconnected terms to the taxonomy (§~\ref{orphan_attach}). In the last step, we further optimize the taxonomy by employing the endocentric nature of hyponyms (§~\ref{remaining_terms}). See Figure \ref{process_chart} for a schematic visualization of the refinement pipeline.
In our experiments, we use the output of three different systems. The refinement method is generically applicable to (noisy) taxonomies, yielding an improved taxonomy extraction system overall.

\subsection{Domain-specific Poincaré Embedding}
\label{Poincare_step}

\paragraph{Training Dataset Construction} 
To create domain-specific Poincaré embeddings, we use noisy hypernym relationships extracted from a combination of general and domain-specific corpora. For the general domain, we extracted 59.2 GB of text from English Wikipedia, Gigaword \citep{parker}, ukWac \citep{Ferraresi} and LCC news corpora \citep{goldhahn}. The domain-specific corpora consist of web pages, selected by using a combination of BootCat \citep{baroni-bernardini-2004-bootcat} and focused crawling \citep{remus2016domain}. Noisy IS-A relations are extracted with lexical-syntactic patterns from all corpora by applying PattaMaika\footnote{\url{http://jobimtext.org}: The PattaMaika component is based on UIMA RUTA \citep{kluegl2016uima}.}, PatternSim \citep{konvens:23_panchenko12p}, and \mbox{WebISA} \citep{Seitner2016ALD}, following \citep{panchenko2016taxi}.\footnote{Alternatively to the relations extracted using lexical patterns, we also tried to use hypernyms extracted using the pre-trained HypeNet model~\cite{shwartz2016improving}, but the overall taxonomy evaluation results were lower than the standard baseline of the TAXI system and thus are not presented here.}

The extracted noisy relationships of the common and domain-specific corpora are further processed separately and combined afterward. To limit the number of terms and relationships, we restrict the IS-A relationships on pairs for which both entities are part of the taxonomy's vocabulary. Relations with a frequency of less than three are removed to filter noise. Besides further removing every reflexive relationship, only the more frequent pair of a symmetric relationship is kept. Hence, the set of cleaned relationships is transformed into being antisymmetric and irreflexive. The same procedure is applied to relationships extracted from the general-domain corpus with a frequency cut-off of five. They are then used to expand the set of relationships created from the domain-specific corpora.

\paragraph{Hypernym-Hyponym Distance} 
Poincaré embeddings are trained on these cleaned IS-A relationships. For comparison, we also trained a model on noun pairs extracted from WordNet (P-WN). Pairs were only kept if both nouns were present in the vocabulary of the taxonomy.
Finally, we trained the word2vec embeddings, connecting compound terms in the training corpus (Wikipedia) by '\emph{\_'} to learn representations for compound terms, i.e multiword units, for the input vocabulary.

In contrast to embeddings in the Euclidean space where the cosine similarity $\frac{\textbf{u} \cdot \textbf{v}}{|\textbf{u}| |\textbf{v}|}$ is commonly applied as a similarity measure, 
Poincaré embeddings use a hyperbolic space, specifically the Poincaré ball model \citep{stillwell1996sources}. Hyperbolic embeddings are designed for modeling hierarchical relationships between words as they explicitly capture the hierarchy between words in the embedding space and are therefore a natural fit for inducing taxonomies. They were also  successfully applied to hierarchical relations in image classification tasks \cite{khrulkov2019hyperbolic}. 
The distance between two points $\textbf{u},\textbf{v} \in \mathcal{B}^d$ for a $d$-dimensional Poincaré Ball model is defined as:
\begin{equation*}
\label{poin_eq}
    d(\textbf{u}, \textbf{v}) = \textrm{arcosh}\Big( 1 + 2 \frac{||\textbf{u} - \textbf{v} ||^2}{(1 - ||\textbf{u}||^2) (1 - ||\textbf{v}||^2)}\Big).
\end{equation*}
This Poincaré distance enables us to capture the hierarchy and similarity between words simultaneously. It increases exponentially with the depth of the hierarchy. So while the distance of a leaf node to most other nodes in the hierarchy is very high, nodes on abstract levels, such as the root, have a comparably small distance to all nodes in the hierarchy.
The word2vec embeddings have no notion of hierarchy and hierarchical relationships cannot be represented with vector offsets across the vocabulary \cite{fu2014learning}. When applying word2vec, we use the observation that distributionally similar words are often co-hyponyms \citep{heylen2008modelling, weeds2014learning}.

\subsection{Relocation of Outlier Terms}
\label{outlier_replace}

Poincaré embeddings are used to compute and store a rank $rank(x,y)$ between every child and parent of the existing taxonomy, defined as the index of $y$ in the list of sorted Poincaré distances of all entities of the taxonomy to $x$.
Hypernym-hyponym relationships with a rank larger than the mean of all ranks are removed, chosen on the basis of tests on the 2015 TExEval data \cite{task17semeval2015}. Disconnected components that have children are re-connected to the most similar parent in the taxonomy or to the taxonomy root if no distributed representation exists. Previously or now disconnected isolated nodes are subject to orphan attachment (§~\ref{orphan_attach}).
 
Since distributional similarity does not capture parent-child relations, the relationships are not registered as parent-child but as co-hyponym relationships. Thus, we compute the distance to the closest co-hyponym (child of the same parent) for every node. This filtering technique is then applied to identify and relocate outliers.

\subsection{Attachment of Orphan Terms}
\label{orphan_attach}
We then attach orphans (nodes unattached in the input or due to the removal of relationships in the previous step) by computing the rank between every orphan and the most similar node in the taxonomy. This node is an orphan's potential parent.
Only hypernym-hyponym relationships with a rank lower or equal to the mean of all stored ranks are added to the taxonomy.
For the word2vec system, a link is added between the parent of the most similar co-hyponym and the orphan.

\subsection{Attachment of Compound Terms}
\label{remaining_terms}
In case a representation for a compound noun term does not exist, we connect it to a term that is a substring of the compound. If no such term exists, the noun remains disconnected. Finally, the Tarjan algorithm \cite{tarjan1972depth} is applied to ensure that the refined taxonomy is asymmetric: In case a circle is detected, one of its links is removed at random.

\begin{figure}[ht]
\mbox{\includegraphics[width = 1\columnwidth, trim={0cm, 0cm, 0cm, 0cm}, clip]{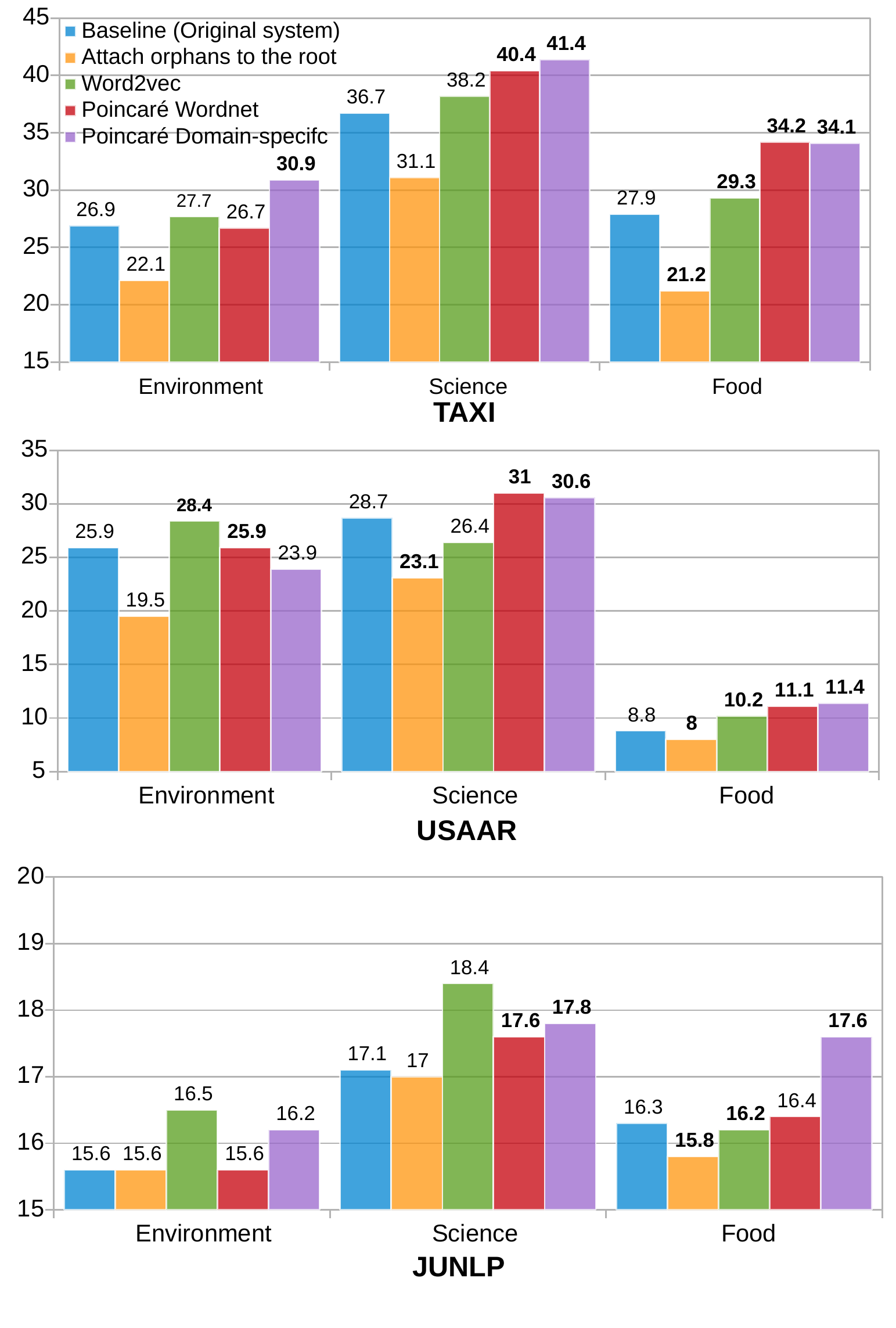}}
\vspace{-2em}
\caption{F$_{1}$ results for the systems on all domains. Vocabulary sizes: environment ($|V| = 261$), science ($|V| = 453$), food ($|V| = 1556$). \textbf{Bold} numbers are significantly different results to the original system with $p < 0.05$.}
\label{results_domains}
\end{figure} 

\section{Evaluation}

Proposed methods are evaluated on the data of SemEval2016 TExEval \cite{bordea2016semeval} for submitted systems that created taxonomies for all domains of the task\footnote{\url{http://alt.qcri.org/semeval2016/task13/index.php}}, namely the task-winning system TAXI \citep{panchenko2016taxi} as well as the systems USAAR \citep{tan2016usaar} and JUNLP \citep{maitra2016junlp}. TAXI harvests hypernyms with substring inclusion and lexical-syntactic patterns by obtaining domain-specific texts via focused web crawling. USAAR and JUNLP heavily rely on rule-based approaches. While USAAR exploits the endocentric nature of hyponyms, JUNLP combines two string inclusion heuristics with semantic relations from BabelNet.
We use the taxonomies created by these systems as our baseline and additionally ensured that taxonomies do neither have circles nor in-going edges to the taxonomy root by applying the Tarjan algorithm \cite{tarjan1972depth}, removing a random link from detected cycles. This causes slight differences between the baseline results in Figure \ref{results_domains} and \cite{bordea2016semeval}.

\section{Results and Discussion}

\paragraph{Comparison to Baselines} 
Figure \ref{results_domains} shows comparative results for all datasets and measures for every system.
The \emph{Root} method, which connects all orphans to the root of the taxonomy, has the highest connectivity but falls behind in scores significantly. 
Word2vec CBOW embeddings partly increase the scores, however, the effect appears to be inconsistent. Word2vec embeddings connect more orphans to the taxonomy (cf. Table \ref{occurence of orphans}), albeit with mixed quality, thus the interpretation of word similarity as co-hyponymy does not seem to be appropriate. Word2vec as a means to detect hypernyms has shown to be rather unsuitable \citep{levy2015supervised}. Even more advanced methods such as the \emph{diff} model \citep{fu2014learning}  merely learn so-called \emph{prototypical hypernyms}.

\begin{table*}[!htbp]
\centering
 \resizebox{1\textwidth}{!}{
\begin{tabular}{lrrrr}
\toprule
Word & Parent in TAXI & Parent after refinement & Gold parent & Closest neighbors\\
\hline
second language acquisition & --- & linguistics & linguistics & applied linguistics, semantics, linguistics\\
botany & --- & genetics & plant science, ecology & genetics, evolutionary ecology, animal science  \\
sweet potatoes & ---& vegetables & vegetables & vegetables, side dishes, fruit\\
wastewater & water & waste  & waste & marine pollution, waste, pollutant\\
water & waste, natural resources & natural resources & aquatic environment & continental shelf, management of resources\\
international relations & sociology, analysis, humanities & humanities & political science & economics, economic theory, geography\\
\bottomrule
\end{tabular}}
\caption{
Example words with respective parent(s) in the input taxonomy and after refinement using our domain-specfic Poincaré embeddings, as well as the word's closest three neighbors (incl. orphans) in embeddings.
 }
\label{most_similar_words}
\end{table*}

\begin{table}[!htbp]
\centering
 \resizebox{1\columnwidth}{!}{
\begin{tabular}{ccccc}
\toprule
Domain & word2vec & P. WordNet & P. domain-specific & \# orphans\\ 
\midrule
Environment & 25 & 18 & 34&  113\\
Science & 56 & 39 & 48 & 158\\
Food & 347 & 181 & 267 & 775 \\
\bottomrule
\end{tabular}}
\caption{
Number of attached orphans in taxonomies created by TAXI using different embeddings. 
}
\label{occurence of orphans}
\vspace{-1em}
\end{table}

Both Poincaré embeddings variants outperform the word2vec ones yielding major improvements over the baseline taxonomy. Employing the \citet{mcnemar1947note} significance test shows that Poincaré embeddings' improvements to the original systems are indeed significant. The achieved improvements are larger on the TAXI system than on the other two systems. We attribute to the differences of these approaches: The rule-based approaches relying on string inclusion as carried out by USAAR and JUNLP are highly similar to step §3.4. Additionally, JUNLP creates taxonomies with many but very noisy relationships, therefore step §3.3 does not yield significant gains, since there are much fewer orphans available to connect to the taxonomy. This problem also affects the USAAR system for the food domain. For the environment domain, however, USAAR creates a taxonomy with very high precision but low recall which makes step §3.2 relatively ineffective. As step §3.3 has shown to improve scores more than §3.2, the gains on JUNLP are comparably lower. 

\paragraph{WordNet-based Embeddings} The domain-specific Poincar\'e embeddings mostly perform either comparably or outperform the WordNet-based ones.
In error analysis, we found that while WordNet-based embeddings are more accurate, they have a lower coverage as seen in Table \ref{occurence of orphans}, especially for attaching complex multiword orphan vocabulary entries that are not contained in WordNet, e.g., \emph{second language acquisition}.
Based on the results we achieved by using domain-specific Poincar\'e embeddings, we hypothesize that their attributes result in a system that learns hierarchical relations between a pair of terms. The closest neighbors of terms in the embedding clearly tend to be more generic as exemplarily shown in Table \ref{most_similar_words}, which further supports our claim. Their use also enables the correction of false relations created by string inclusion heuristics as seen with \emph{wastewater}. However, we also notice that few and inaccurate relations for some words results in imprecise word representations such as for \emph{botany}. 

\paragraph{Multilingual Results} Applying domain-specific Poincar\'e embeddings to other languages also creates overall improved taxonomies, however the scores vary as seen in Table \ref{multi-language results}. While the score of all food taxonomies increased substantially, the taxonomies quality for environment did not improve, it even declines. This seems to be due to the lack of extracted relations in (§3.1), which results in imprecise representations and a highly limited vocabulary in the Poincar\'e embedding model, especially for Italian and Dutch. In these cases, the refinement is mostly defined by step §3.4.

\begin{table}[!htbp]
\centering
 \resizebox{1\columnwidth}{!}{
\begin{tabular}{cccccc}
\toprule
Language & Domain & Original & Refined & \# rel. data & \# rel. gold\\
\hline
English & Environment & 26.9 & \textbf{30.9} & 657 & 261\\
& Science & 36.7 & \textbf{41.4} & 451 & 465\\
& Food & 27.9 & \textbf{34.1} & 1898 & 1587\\
\midrule
French & Environment & 23.7 & \textbf{28.3} & 114 & 266\\
& Science & 31.8 & \textbf{33.1} & 118 & 451\\
& Food & 22.4 & \textbf{28.9} & 598 & 1441\\
\hline
Italian & Environment & \textbf{31.0} & 30.8 & 2 & 266\\
& Science & 32.0 & \textbf{34.2} & 4 & 444\\
& Food & 16.9 & \textbf{18.5} & 57 & 1304\\
\hline
Dutch & Environment & \textbf{28.4} & 27.1 & 7 & 267\\
& Science & 29.8 & \textbf{30.5} & 15 & 449\\
& Food & 19.4 & \textbf{21.8} & 61 & 1446\\
\bottomrule
\end{tabular}}
\caption{
F$_{1}$ comparison between original (TAXI) and refined taxonomy using domain-specific embeddings. 
}
\label{multi-language results}  
\end{table}

\section{Conclusion}

We presented a refinement method for improving existing taxonomies through the use of hyperbolic Poincar\'e embeddings. They consistently yield improvements over strong baselines and in comparison to word2vec as a representative for distributional vectors in the Euclidean space.
We further showed that Poincar\'e embeddings can be efficiently created for a specific domain from crawled text without the need for an existing database such as WordNet.
This observation confirms the theoretical capability of Poincar\'e embeddings to learn hierarchical relations, which enables their future use in a wide range of semantic tasks. 
A prominent direction for future work is using the hyperbolic embeddings as the sole signal for taxonomy extraction. Since distributional and hyperbolic embeddings cover different relations between terms, it may be interesting to combine them.

\subsubsection*{Acknowledgments}

We acknowledge the support of DFG under the ``JOIN-T'' (BI 1544/4) and ``ACQuA'' (BI 1544/7) projects as well as the DAAD. We also thank three anonymous reviewers and Simone Paolo Ponzetto for providing useful feedback on this work.

\bibliographystyle{acl_natbib.bst}
\bibliography{bibliography.bib}

\end{document}